# Intelligent Multi-channel Meta-imagers for Accelerating Machine Vision

Hanyu Zheng[1], Quan Liu[2], Ivan I. Kravchenko[3], Xiaomeng Zhang[4], Yuankai Huo[2], and Jason G. Valentine[4*].

1. Department of Electrical and Computer Engineering, Vanderbilt University, Nashville, TN, USA, 37212.
2. Department of Computer Science, Vanderbilt University, Nashville, TN, USA, 37212.
3. Center for Nanophase Materials Sciences, Oak Ridge National Laboratory, Oak Ridge, TN, USA, 37830.
4. Department of Mechanical Engineering, Vanderbilt University, Nashville, TN, USA, 37212.

* Corresponding author: jason.g.valentine@vanderbilt.edu

**Abstract**

Rapid developments in machine vision have led to advances in a variety of industries, from medical image analysis to autonomous systems. These achievements, however, typically necessitate digital neural networks with heavy computational requirements, which are limited by high energy consumption and further hinder real-time decision-making when computation resources are not accessible. Here, we demonstrate an intelligent meta-imager that is designed to work in concert with a digital back-end to off-load computationally expensive convolution operations into high-speed and low-power optics. In this architecture, metasurfaces enable both angle and polarization multiplexing to create multiple information channels that perform positive and negatively valued convolution operations in a single shot. The meta-imager is employed for object classification, experimentally achieving 98.6% accurate classification of handwritten digits and 88.8% accuracy in classifying fashion images. With compactness, high speed, and low power consumption, this approach could find a wide range of applications in artificial intelligence and machine vision applications.

# Introduction

The rapid development of digital neural networks and the availability of large training datasets have enabled a wide range of machine-learning-based applications, including image analysis[1,2], speech recognition[3,4], and machine vision[5]. However, enhanced performance is typically associated with a rise in model complexity, leading to larger compute requirements[6]. The escalating use and complexity of neural networks have resulted in increases in energy consumption while limiting real-time decision-making when large computational resources are not readily accessible. These issues are especially critical to the performance of machine vision[7,8,9] in autonomous systems where the imager and processesor must have small size, weight, and power consumption for on-board processing while still maintaining low latency, high accuracy, and highly robust operation. These opposing requirements necessitate the development of new hardware and software solutions as the demands on machine vision systems continue to grow.

Optics has long been studied as a way to speed computational operations while also increasing energy efficiency[10,11,12,13,14,15]. In accelerating vision systems there is the unique opportunity to off-load computation into the front-end imaging optics by designing an intelligent imager that is optimized for a particular computational task. Free-space optical computational, based on Fourier optics[16,17,18,19], actually predates modern digital circuitry and allows for highly parallel execution of the convolution operations which comprise the majority of the floating point operations in machine vision architectures[20,21]. The challenge with Fourier-based processors is that they are traditionally employed by reprojecting the imagery using spatial light modulators and coherent sources, enlarging the system size compared to chip-based approaches[22,23,24,25,26,27]. While coherent illumation is not strictly required, it allows for more freedom in the convolution operations including the ability to achieve the negatively valued kernels needed for spatial derivatives. Optical diffractive neural networks[28,29,30] offer an alternative approach though these are also employed with coherent sources and thus are best suited as back-end processors with image data being reprojected.

Metasurfaces offer a uniqe platform for implementing front-end optical computation as they can reduce the size of the optical elements while allowing for a wider range of optical properties including polarization[31,32], wavelength[33], and angle of incidence[34,35] to be utilized in computation. For instance, metasurfaces have been demonstrated with angle of incidence dependent transfer functions for realizing compact optical differentiation systems[36,37,38,39] with no need to pass through the Fourier plane of a two lens system. In addition, wavelength multiplexed metasurfaces, combined with optoelectronic subtraction, have be used to achieve negatively valued kernels for excecuting single-shot differentiation with incoherent light[40,41]. Differentiatoin, however, is a single convolution operation while most machine vision systems require multiple independent channels. There has been recent work on multi-channel convolutional front-ends but these have been limited in transmission efficiency and computational complexity, achieving only positively valued kernels with a stride that is equal to the kernal size, preventing implementation of common digital designs[42,43]. While these are important steps towards a computational front-end, an architecture is still needed for generating the multiple independent, and arbitrary, convolution channels that are used in machine vision systems.

Here, we demonstrate an intelligent meta-imager that can serve as a multi-channel convolutional accelerator for incoherent light. To achieve this, the point spread function (PSF) of the imaging meta-optic is engineered to achieve parallel multi-channel convolution using a single aperture implemented with angular multiplexing, as shown in Fig.1. In addition, positively and negatively valued kernels are achieved for incoherent illumination by using polarization multiplexing[44], combined with a polarization-sensitive camera and optoelectronic subtraction. A second metasurface corrector is also employed to widen the field of view (FOV) for imaging objects in the natural world and both metasurfaces are restricted to phase functions, yielding high transmission efficiency. As a proof-of-concept, the platform is used to experimentally demonstrate classification of the MNIST and Fashion-MNIST datasets[45] with measured accuracies of 98.6% and 88.8%, respectively. In both cases, 94% of the FLOPs are off-loaded from the digital platform into the front-end optics. Due to the compact footprint, freedom in kernel design, and ability to generate multiple independent channels, this type of meta-imager is suitable for use in a wide variety of neural network architectures used in machine vision including convolutional neural networks (CNNs), image processors, and image transformers.

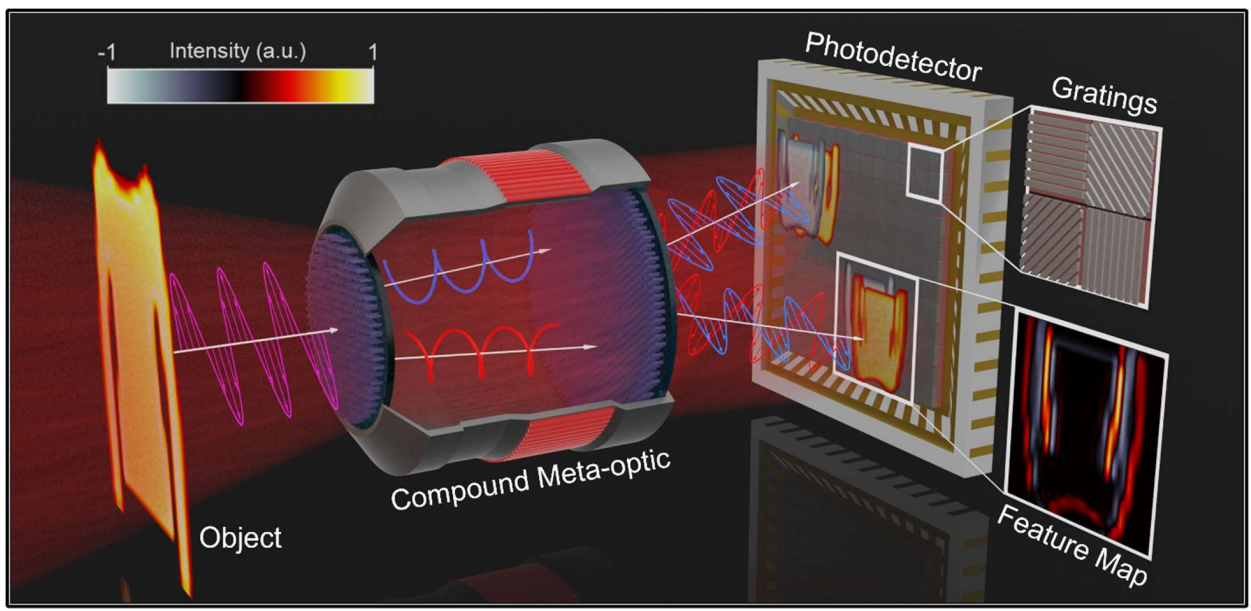

**Figure 1. Schematic of the meta-imager.** The meta-imager enables multi-channel signal processing for replacing convolution operations in a digital neural network. Positive and negative values are distinguished and recorded as feature maps by a polarization-sensitive photodetector.

**Results and Discussion**

The meta-optic described here is designed to optically implement the convolutional layers at the front-end of a digital neural network. In a digital network, convolution comprises matrix multiplication of the object image and an $N \times N$ pixel kernel with each pixel having an independent weight, as illustrated for the case of $N = 3$ in Fig. 2 (a). The kernel is multiplied over an area of the image using a dot product and then rastered across the image, moving by a single pixel each step until it is swept across the entire image, forming a single feature map. Under incoherent illumination, optical convolutional is expressed as $Image = Object \otimes PSF(x, y)$ where $PSF(x, y)$ is the point spread function of the optic. Typically, in implementing the optical version of digital convolution the $PSF(x, y)$ is the continuous function that was discretized in forming the digital kernel. Here, we take a different approach, creating a true optical analog to the digital kernel.

This is done by engineering the $PSF(x,y)$, as shown in Fig. 2(a), to possess $N \times N$ focal spots, each with a different weight, or image intensity, that matches the desired digital kernel weight. These focal spots will result in $N \times N$ images of the object being formed that are spatially overlapped on the sensor and offset based on the separation in the focal spot positions. In this case, we are rastering weighted images with the summing operation in the dot product being achieved by overlapping the images on the camera.

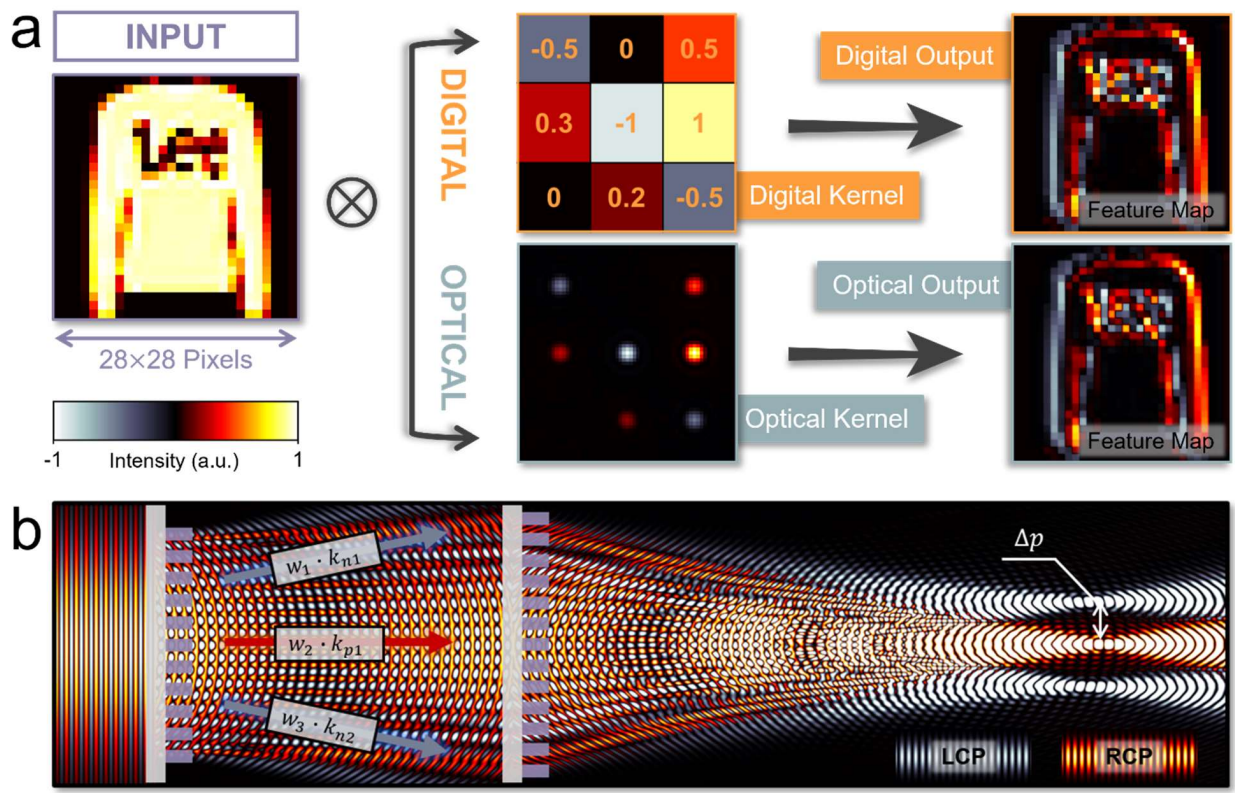

**Figure 2. Meta-optic Architecture.** (a) Comparison between the digital and optical convolution process. A random $3 \times 3$ kernel, normalized between $[-1,1]$, was defined to convolve an image digitally. The equivalent optical PSF was designed and simulated by the angular spectrum propagation method, with the optical output calculated based on the premise of a coma-free system. (b) The architecture of the compound meta-optic forms three independent focal spots as the PSF. Angular multiplexing is used in the first layer metasurface, which can split light into multiple signal channels and correct the wavefront for wide-view-angle imaging. Meanwhile, polarization multiplexing is used to realize an independent response for orthogonal polarization states. In our case right-hand-circular (RCP) and left-hand-circular (LCP) polarized signals are used for positive and negative kernel values, respectively.

In this architecture, positively and negatively valued kernel weights are achieved by encoding the focal spots with either right-hand-circular polarization (RCP) or left-hand-circular (LCP), respectively. The RCP and LCP feature maps, shown in Fig. 2(a), are independently recorded using a camera with polarization filters with summing being achieved by digitally subtracting the LCP feature map from the RCP feature map. The convolution generated by this method is identical to the digital process which is evidenced by comparing the digital and optical feature maps in Fig. 2(a). We have used this approach for several reasons. First, as will be explained, the phase and amplitude profile associated with our desired $PSF(x,y)$ is analytical, significantly simplifying the design process and allowing us to achieve numerous independent feature maps, or channels, using one aperture. In addition, since we have a true optical analog to a digital system, we can directly implement digital kernel designs with optics, removing the optic from the design loop, further speeding the design process. In order to achieve the desired optical response, we employ a bilayer metasurfaces architecture, as shown in Fig. 2(b). In this architecture, the first metasurface splits the incident signal into angular channels of varying weight while birefringence in this layer is used to encode positive and negative kernel values in RCP and LCP polarization, respectively. The second metasurface is polarization insensitive and serves as the focusing optic to create a $N$ x $N$ focal spot array for each channel.

Meta-optic design began by optimizing a two metasurface lens, comprising a wavefront corrector and focuser, to be coma-free over a ±10° angular range using the commercial software, Zemax (see details in methods). The angular response of the metasurfaces shows constant focal spot shape within the designed angular range. Once the coma-free meta-optic was designed, angular multiplexing was applied to the first metasurface to form focal spot arrays as the convolution kernels. The focal spot position is controlled using angular multiplexing with each angle corresponding to a kernel pixel. By encoding a weight to each angular component, the system PSF, serving as the optical kernel, can be readily engineered. The analytical expression of the complex-amplitude profile multiplexing all angular signals is given by,

$$A(x,y) = \sum_{m}^{M}\sum_{n}^{N} \sqrt{w_{mn}} \exp\left\{i\frac{2\pi}{\lambda}\left[x\sin(\theta_{x|mn}) + y\sin(\theta_{y|mn})\right]\right\} \qquad (1)$$

where $A(x,y)$ is a complex-amplitude field. $M, N$ is the row and column number of elements in the kernel. $w_{mn}$ is the corresponding weight of each element, which is normalized to a range of [0,1]. $\lambda$ is the working wavelength, $x$ and $y$ are the spatial coordinates, and $\theta_{x|mn}$ and $\theta_{y|mn}$ are the designed angles with a small variation to form the kernel elements. The deflection angles are selected to realize the desired PSF for incoherent light illumination which is given by,

$$PSF(x,y) = \sum_{m}^{M}\sum_{n}^{N} w_{mn} \Theta\left\{x - f_1 c\left[\frac{x_0}{f_2} + \tan(\theta_{x|mn})\right], y - f_1 c\left[\frac{y_0}{f_2} + \tan(\theta_{y|mn})\right]\right\} \qquad (2)$$

where $x_0$ and $y_0$ are the location of the object and $\Theta(x,y)$ is the focal spot excited by a plane wave. $f_1$ is focal length of the meta-imager while $c$ is a constant fitted based on the imaging system. $f_2$ is the distance from the object to the front aperture. The separation distance of each focal spot, $\Delta p$, defines the imaged pixel size of the object. Based on a prescribed PSF the required angles, $\theta_{mn}$, can be derived from Eq.2, which can be further extended into an off-axis imaging case for the purpose of multi-channel, single-shot convolutional applications.

In Eq.1 we employ a spatially varying complex-valued amplitude function that would ultimately introduce large reflection loss leading to a low diffraction efficiency[46]. To overcome this limitation, an optimization platform was developed based on the angular spectrum propagation method and stochastic gradient descent (SGD) solver, which converts the complex-amplitude profile into a phase-only metasurface. The algorithm encodes a phase term, $\exp(i\phi_{mn})$, onto each weight, $w_{mn}$, based on the loss function, $\mathcal{L} = \sum(|A|^2 - I)^2/N$. Here, $I$ is a matrix consisting of unity elements and $N$ is the total pixel number. The intensity profile becomes more consistent and closer to a phase-only device by minimizing the loss function during optimization. The phase-only approximation can effectively avoid loss in the complex-amplitude function, leading to a theoretical diffraction efficiency as high as 84.3% where 14% of the loss is introduced by Fresnel reflection, which can be removed by adding anti-reflection coatings.

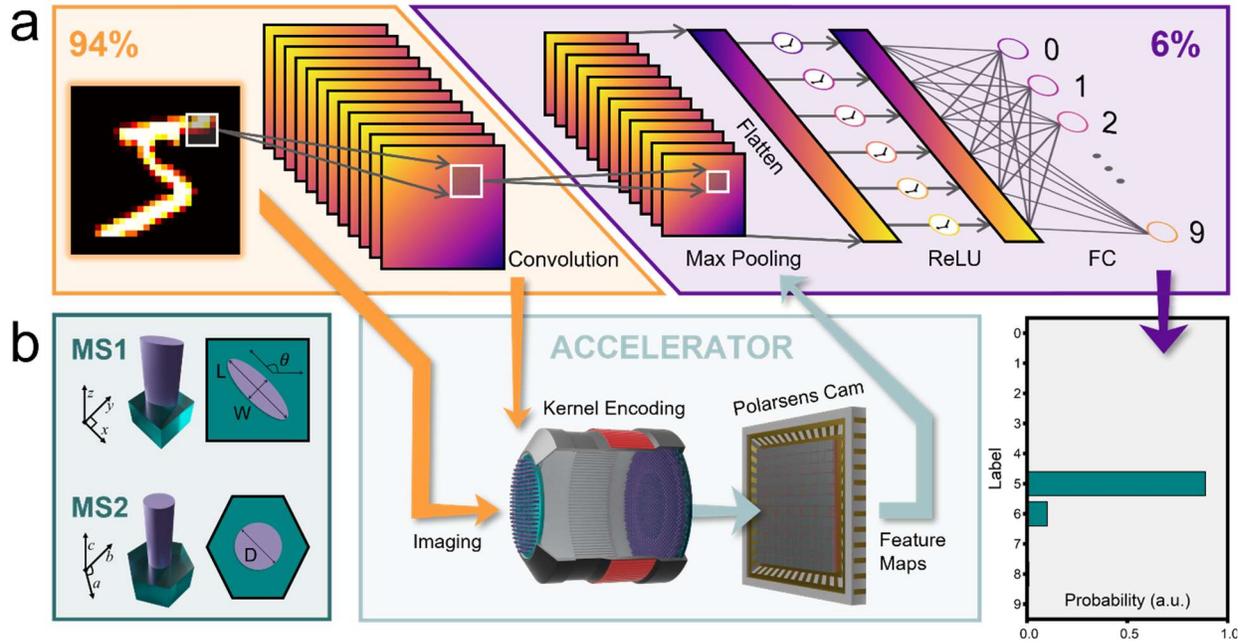

**Figure 3. Design of the Meta-imager.** (a) Design process of the hybrid neural network. A shallow convolutional neural network was trained at first. In this case, the input is convoluted by 12 independent channels, each comprising 7×7 pixel kernels. The convolution operations are implemented using the meta-imager, with the extracted feature maps, including multiplexed polarization channels, recorded by a polarization-sensitive camera. The processed feature maps were then fed into the pre-trained digital neural network to obtain the probability histogram for image classification. The number at the corner indicates the percentage of relevant computing operations in total FLOPs. (b) The schematic of the meta-atoms for the first and second metasurfaces. The height is fixed at 0.6μm while the lattice constant is chosen as 0.45μm and 0.47μm, respectively.

In order to validate the performance of this architecture, a shallow CNN was trained for the purpose of image classification. The neural network architecture, shown in Fig.3 (a), contains an optical convolution layer followed by digital max pooling, a rectified linear (ReLU) activation function, and a fully connected (FC) layer. In the convolution process, 12 independent kernels are used to extract feature maps and the overall intensity of positive and negative channels was set to be equal due to energy conservation from the phase-only approximation in the meta-optic design. Since neural network training is a high-dimensional problem with infinite solutions, the above kernel restrictions do not significantly affect the final performance. Each kernel comprised $N = 7$ pixels instead of a more typical $N = 3$ format, to correlate neurons within a broader viewing field[47], leading to better performance for large-scale object recognition. The detailed training process is described in the methods section. In order to finish classification, the feature maps extracted by the compound meta-optic are fed into the digital component of the neural network. In this architecture 94% of the total FLOPs are off-loaded from the digital platform into the meta-optic leading to a significant speedup for classification tasks.

To realize the first, polarization selective metasurface, elliptical nanopillars were chosen as the base meta-atoms, as shown in Fig.3 (b). The width and length of nanopillars were designed so that the nanopillars serve as half-wave plates. This choice introduces spin-decoupled phase response by introducing geometrical and a locally-resonant phase delay simultaneously, hence independent phase control over orthogonal circular-polarized states can be achieved. The analytical expression of the phase delay for the different polarization states is described as,

$$\begin{bmatrix}\phi_{LCP}\\ \phi_{RCP}\end{bmatrix} = \begin{bmatrix}\phi_x + 2\theta + \pi/4\\ \phi_x - 2\theta - \pi/4\end{bmatrix} \quad (3)$$

Here, $\phi_x$ is the phase delay of the meta-atoms along $x$ axis at $\theta = 0$. Hence, by tuning the length, width, and rotation angle, the phase delay of LCP and RCP light can be independently controlled. The second metasurface was designed based on circular nanopillars arranged in a hexagonal lattice for realizing polarization-insensitive phase control.

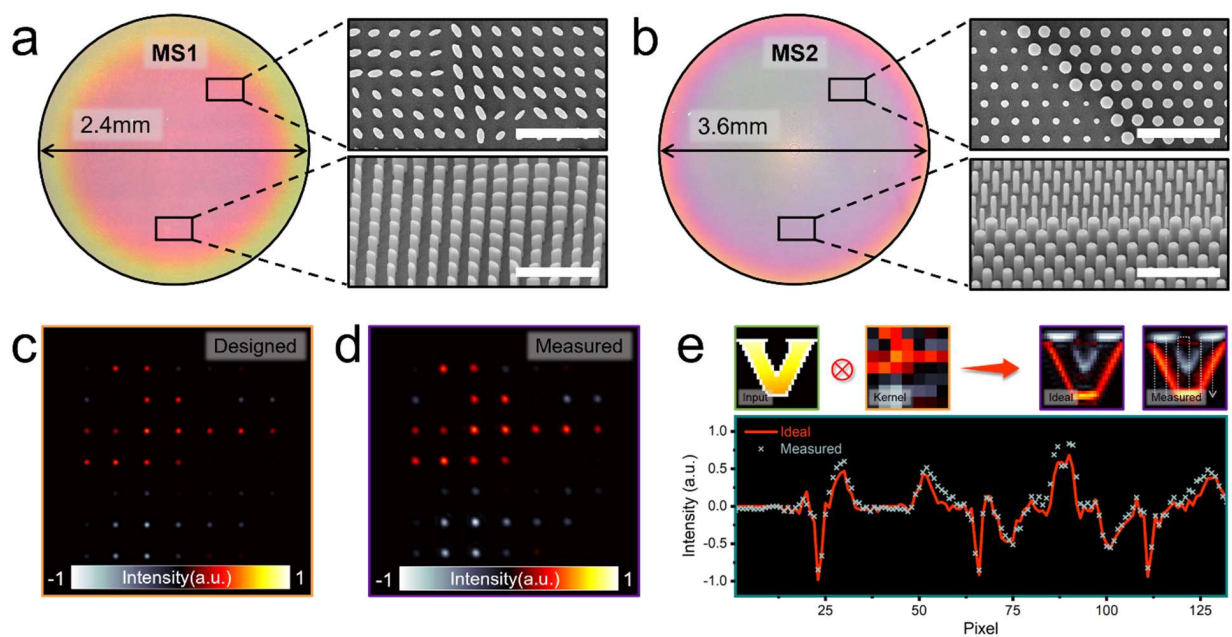

**Figure 4. Fabrication and Characterization of the Meta-imager.** (a) and (b) Optical images of the fabricated metasurfaces comprising the meta-imager. The inset is an SEM image of each metasurface. Scale bar: 5μm. (c) An ideal optical kernel calculated based on the angular spectrum propagation method. The weight of each spot is equal to the pre-designed digital kernel. (d) Measured intensity profile of the kernel generated by the fabricated meta-optic. (e) Comparison between convolutional results based on the ideal and measured kernels. The solid white line indicates the sampled pixels for comparison. The demonstration kernel is the same as (c) and (d).

Two versions of the meta-optic classifier were fabricated based on networks trained for MNIST and Fashion-MNIST datasets. Fabrication of the meta-optic began with a silicon device layer on a fused silica substrate patterned by the standard electron beam lithography (EBL) followed by reactive-ion-etching (RIE). A thin polymethyl methacrylate (PMMA) layer was spin-coated over the device as the protective and index-matching layer. The detailed fabrication process is described in the Methods section. An optical image of the two metasurfaces comprising the meta-optic is exhibited in Fig.4 (a) and (b) with the inset showing the meta-atoms. In order to align the compound meta-optic, the first metasurface was mounted in a rotational stage (CRM1PT, Thorlabs) while the second layer was fitted in a 3-axis translational stage (CXYZ05A, Thorlabs). The metasurfaces are aligned in situ and characterized in a cage system. A meta-hologram was fabricated on the first layer alongside the device to assist the alignment process to form an alignment pattern at a prescribed distance along the optical axis corresponding to the designed separation distance. The alignment process was finished by overlapping the alignment pattern with the low-transmission register on the second layer. Due to the large size (mm-scale) of each metasurface layer, the meta-optic exhibits high alignment tolerance. The system performance remains constant under a horizontal misalignment of 65μm and vertical displacement of ±400μm, indicating the robustness of the entire convolutional system.

In order to characterize the optical properties of the fabricated meta-optic, a linearly polarized laser was used for illumination in obtaining the PSF. The linearly polarized light source includes LCP and RCP components with equal strength. The PSF at the focal plane of the compound meta-optic, shown in Fig.4 (c) and (d), indicates a good match between the ideal and measured results, where the red and blue represent positive and negative values, respectively.

Optical convolution of a grayscale Vanderbilt logo was used to characterize the accuracy of the fabricated meta-optic, as shown in Fig.4 (e). To accomplish this, an imaging system using a liquid-crystal-based spatial light modulator (SLM) was built. An incoherent tungsten lamp with a bandpass filter was used for SLM illumination. The feature maps extracted by the meta-optic were recorded by a polarization-sensitive camera (DZK 33UX250, Imaging Source) where orthogonally polarized channels are simultaneously recorded using polarization filters on each camera pixel. The comparison between the digital and measured feature maps, recorded on the camera, is illustrated in Fig.4 (e). The pixel intensity from digital and measured convolutional results at the same position were extracted and compared to evaluate the convolution fidelity. The deviation between the ideal and measured results, defined by $\sigma = \sum_{n=1}^{N}|D_{i,n} - D_{m,n}|/(2N)$, was calculated as 3.83%, where $D_i$ and $D_m$ are the ideal and measured intensity, and $N$ is the number of total pixels. The error originates from stray light, fabrication imperfections, the local phase approximation, and metasurface misalignment. These errors also result in a small amount of zeroth order diffracted light being introduced from the first metasurface leading to a spot at the center of the imaging plane. However, the polarization state of the zeroth order light remains unchanged, with the energy evenly distributed in the two circular polarized channels. Hence, subtraction between the information channels allows the zeroth order pattern to be canceled, not affecting the classification performance.

As a proof-of-concept in demonstrating multi-channel convolution, a full meta-optic classifier was first designed and fabricated based on classification of the MNIST dataset, which includes 60,000 28x28 pixel hand-written digit training images. The feature maps of 1000 digits, not in the training set, were extracted using the meta-optic to characterize the system performance. An example input image is exhibited in Fig.5 (a), with the corresponding feature maps shown in Fig.5 (b). The measured feature maps match well with the theoretical prediction, as shown in Fig.5 (b), indicating good fidelity in the optical convolution process. The theoretical and experimental confusion matrices for this testing dataset are shown in Fig.5 (c), demonstrating 99.3% accurate classification in theory and 98.6% accurate classification in the measurement. The small drop in accuracy likely results from the small inaccuracy in the realized optical kernels.

In order to explore the flexibility of the approach a dataset with higher spatial frequency information, Fashion-MNIST, was also used for training the model with an example input image provided in Fig.5 (d). This dataset includes 60,000 training images of clothing articles that contain images with higher spatial frequencies than the MNIST handwritten digit dataset. The ideal and measured feature maps are compared in Fig.5 (e), indicating good agreement. The confusion matrices for Fashion-MNIST are illustrated in Fig.5 (f), with 90.2% accurate classification in theory and 88.8% in measurement. Compared to the MNIST dataset, the Fashion-MNIST model has a slightly lower accuracy, compared to theory, due to the higher resolution features in the dataset. Specifically, for class 7 in the Fashion-MNIST dataset, the accuracy predicted by the optical frontend dropped from 81.4% to 67.0%, with the model miss-identifying the images as classes 1,3,4,5. We expect these classes to share the same features during model training. These mixed features can be potentially distinguished by adaptively tuning the loss function during model training[48] or utilizing novel neural network architecture such as vision transformer[49] (ViT) with better performance at comparable FLOPs.

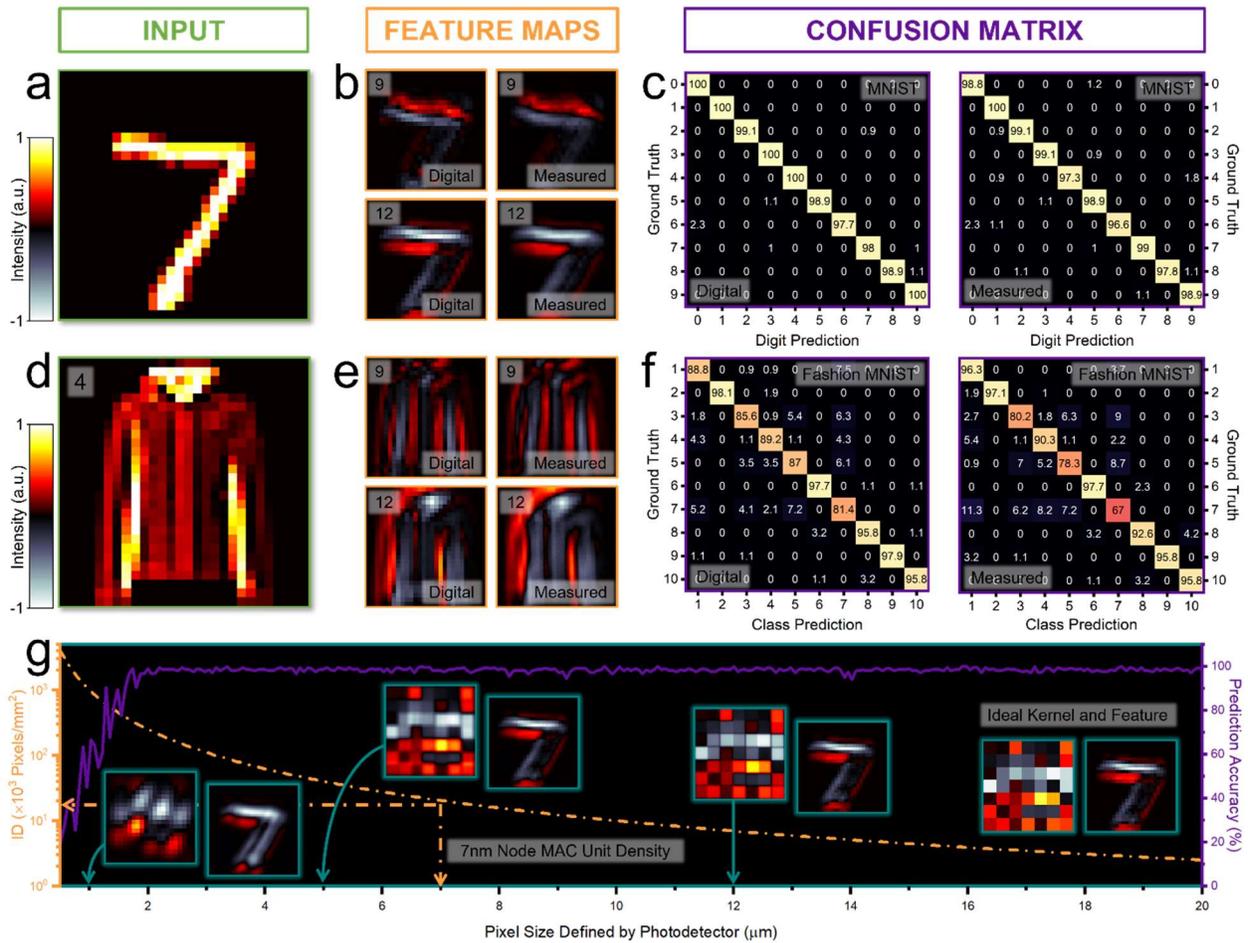

**Figure 5. Classification of MNIST and Fashion-MNIST objects.** (a) An input image from the MNIST dataset. (b) Ideal and experimentally measured feature maps corresponding to the convolution of (a) with channels 1 and 4. The upper-left corner label indicates the channel number during convolution. (c) Comparison between the theoretical and measured confusion matrices for MNIST classification. (d) An input image from the Fashion-MNIST dataset (e) Ideal and experimentally measured feature maps corresponding to the convolution of (a) with channels 1 and 4. The upper-left corner label indicates the channel number during convolution. (f) Comparison between the theoretical and measured confusion matrices for Fashion-MNIST classification. (g) Predicted accuracy curve for the MNIST dataset and information density (ID) as a function of pixel size. The insets depict kernel profiles and feature maps at different pixel sizes.

To understand the computational capacity of the meta-optic, the information density (ID), defined by pixels per unit area, was calculated as a function of the designated pixel size, as shown in Fig.5 (g). Here, the pixel size for image analysis is dictated by the separation distance between the neighboring focal spots in the PSF, which is ultimately dictated by the diffraction limit. The prediction accuracy based on the MNIST dataset in terms of pixel size was computed to benchmark the meta-optic accelerator. The theoretical accuracy remains as high as ~99% until the pixel size drops to 2μm, at which point neighboring focal spots are below the diffraction limit, resulting in additional aberration in the output features. Above the 2μm limitation, the PSF and extracted feature maps remain almost identical to the ideal prediction, as shown in Fig.5 (g) insets. Although a pixel size of 12μm is demonstrated in this work for the proof of concept, the system functionality would remain unchanged in theory when further decreasing the size of the PSF lattice. As a result, the ID of our proposed meta-optic system can go well beyond state-of-the-art multiply-accumulation (MAC) unit density based on the current 7nm node architecture[50].

**Conclusion**

While our proof-of-concept demonstration uses low-resolution images, the large achievable information density and the parallel nature of the optical operations indicate that this architecture could be a key tool for high-speed and low-power machine vision applications. Realizing negatively valued kernels allows for arbitrary convolution operations while angular multiplexing results in a direct analog to digital implementations. Furthermore, the architecture is designed for incoherent illumination and a reasonably wide FOV, both of which are needed for implementation in imaging natural scenes with ambient illumination. Although a tradeoff exists between the channel number and the viewing angle range, a multi-aperture architecture could be designed without deteriorating the FOV in a single imaging channel[51]. As a result, these intelligent meta-imagers can be highly parallel and bridge the gap between the natural world and digital neural networks, allowing one to harness the advantages of both free-space and chip-based architectures. Due to the compact footprint, ease of integration with conventional imaging systems, and ability to access additional information channels, this type of system could find uses in machine vision for autonomous systems[52], information security[53, 54], and quantum communications[55].

**Methods**

*Optimization of Coma-free Meta-optic.* The coma-free meta-optic contains two metasurfaces, whose phase profiles were optimized by the ray tracing technique using commercial optical design software (Zemax OpticSutdio, Zemax LLC). The phase profile of each layer was defined by even order polynomials according to the radial coordinate, $\rho$, as follows:

$$\phi(\rho) = \sum_{n=1}^{5} a_n \left(\frac{\rho}{R}\right)^{2n} \tag{1}$$

where $R$ is the radius of the metasurface, and $a_n$ is the optimized coefficient to minimize the focal spot size of the bilayer metasurfaces system under an incident angle up to $13°$. The diameter of the second layer metasurface was 1.5 times that of the first layer to capture all light under high incident angle illumination. The phase profiles were then wrapped within 0 to $2\pi$ to be fitted by meta-atoms.

*Digital Neural Network Training.* The MNIST and Fashion-MNIST database, each containing 60,000 28x28 pixel training images, were used to train the digital convolutional neural network. The channel number for convolution was set to 12, while the kernel size was fixed at 7×7, with the size of the convolutional result remaining the same. The details of neural network architecture are shown in Fig.3 (a) in the main context. During forward propagation in the neural network, an additional loss function defined by $\mathcal{L} = \sum_{n=1}^{N} w_n$ was added to ensure equal total intensity of positive and negative kernel values, where $w_n$ is the weight of each kernel. All the kernel values are normalized to $[-1,1]$, by dividing by a constant, to maximize the diffraction efficiency in the optics. An Adam optimizer was utilized for training the digital parameters with a learning rate of 0.001. The training process is sustained over 50 epochs, during which the performance is optimized by minimizing the negative log-likelihood loss from comparing prediction probabilities and ground truth labels. The algorithm was programmed based on Pytorch 1.10.1 and CUDA 11.0 with a Quadro RTX 5000/PCIe/SSE2 as the graphics cards.

*Numerical Simulation.* The complex transmission coefficients of the silicon nanopillars were calculated using an open-source rigorous coupled wave analysis (RCWA) solver, Reticolo[56]. A square lattice with a period of 0.45μm was used for the first metasurface with the working wavelength at 0.87μm. The second metasurface was assigned a hexagonal lattice with a period of 0.47μm. During full-wave simulation, the index of silicon and fused silica characterized by ellipsometry was set at 3.74 and 1.45, respectively.

*Metasurface Fabrication.* EBL-based lithography was used to fabricate all the metasurface layers. First, low-pressure chemical vapor deposition (LPCVD) was utilized to deposit a 630nm thick silicon device layer on a fused silica substrate. PMMA photoresist was then spin-coated on the silicon layer, followed by thermal evaporation of a 10nm thick Cr conduction layer. The EBL system then exposed the photoresist, and after removing the Cr layer, the pattern was developed by the MIBK/IPA solution. A 30nm $Al_2O_3$ hard mask was deposited via electron beam evaporation, followed by a lift-off process with N-methyl-2-pyrrolidone (NMP) solution. The silicon was then patterned using reactive ion etching, and a 1μm thick layer of PMMA was spin-coated to encase the nanopillar structures as a protective and index-matching layer.

**Author Contributions**

**Data Availability**

All data needed to evaluate the conclusions in the paper are present in the paper and/or the Supplementary Materials.

## Acknowledgments

HZ and JGV acknowledge support from DARPA under contract HR001118C0015 and NAVAIR under contract N6893622C0030. XZ acknowledges support from ONR under contract N000142112468. YH and QL acknowledge support from NIH under contract R01DK135597.

## Competing Interest

The authors declare no competing interests.